\documentclass{article}
\usepackage{PRIMEarxiv}
\usepackage[utf8]{inputenc} % allow utf-8 input
\usepackage[T1]{fontenc}    % use 8-bit T1 fonts
\usepackage{hyperref}       % hyperlinks
\usepackage{url}            % simple URL typesetting
\usepackage{booktabs}       % professional-quality tables
\usepackage{amsfonts}       % blackboard math symbols
\usepackage{nicefrac}       % compact symbols for 1/2, etc.
\usepackage{microtype}      % microtypography
\usepackage{lipsum}
\usepackage{fancyhdr}
\usepackage{amsmath,amssymb,mathtools}
\usepackage{graphicx,float}
\usepackage{subfig}
\newtheorem{theorem}{Theorem}
\usepackage{bm} 
\usepackage{xcolor}
\hypersetup{
    colorlinks,
    linkcolor={red!50!black},
    citecolor={blue!50!black},
    urlcolor={blue!80!black}
}

\title{Inductive Conformal Prediction: A Straightforward Introduction with Examples in Python
\thanks{\textit{\underline{Citation}}: 
\textbf{Sousa, Martim . Inductive Conformal Prediction: A Straightforward Introduction with Examples in Python.}} 
}

\author{
  Martim Sousa \\
  IEETA/DETI \\
  University of Aveiro \\
  Aveiro\\
  \texttt{martimsousa@ua.pt} \\
}

\begin{document}
\maketitle

\begin{abstract}
Inductive Conformal Prediction (ICP) is a set of distribution-free and model agnostic algorithms devised to predict with a user-defined confidence with coverage guarantee. Instead of having \textit{point predictions}, i.e., a real number in the case of regression or a single class in multi class classification, models calibrated using ICP output an interval or a set of classes, respectively. ICP takes special importance in high-risk settings where we want the true output to belong to the prediction set with high probability. As an example, a classification model might output that given a magnetic resonance image a patient has no latent diseases to report. However, this model output was based on the most likely class, the second most likely class might tell that the patient has a 15\% chance of brain tumor or other severe disease and therefore further exams should be conducted. Using ICP is therefore way more informative and we believe that should be the standard way of producing forecasts. This paper is a hands-on introduction, this means that we will provide examples as we introduce the theory.
\end{abstract}

% keywords can be removed
\keywords{Conformal prediction \and Confidence intervals \and Confidence sets \and Quantile Regression \and Tutorial}

\section{Crash Course on Inductive Conformal Prediction}
Before introducing the general outline of ICP, we want to make it clear that ICP requires data splitting; however, there are other computationally onerous approaches that do not require splitting such as full conformal prediction \cite{1}. Suppose a dataset $D=\{(x_i,y_i)\}_{i=1}^N$, where $x_i \in \mathbb{R}^D$ are the features and $y_i\in \mathbb{R}^Q$ the response variable. As we said, the first step is to split D in three mutually exclusive sets: (i) a training, (ii) a calibration set, and (iii) a validation set such that $D=D_{train} \cup D_{cal} \cup D_{val}$ with $N=n_{train}+n_{cal}+n_{val}$. The general outline of ICP is as follows.
\begin{enumerate}
    \item Train a Machine Learning (ML) model on the training set $D_{train}=\{(x_i,y_i)\}_{i=1}^{n_{train}}$;
    \item Define a heuristic notion of uncertainty given by $s(x,y)$, often referred to as the \textit{non-conformity score function};
    \item For each element $(x,y)\in D_{cal}=\{(x_i,y_i)\}_{i=1}^{n_{cal}}$ apply the function s to get $n_{cal}$ \textit{non-conformity scores} $\{(s_i)\}_{i=1}^{n_{cal}}$;
    \item Select a user-defined miscoverage error rate $\alpha$ and compute $\hat{q}$ as the $\frac{\left \lceil (n_{cal}+1)(1-\alpha)\right \rceil}{n_{cal}}$ quantile of the \textit{non-conformity scores} $\{(s_i)\}_{i=1}^{n_{cal}}$;
    \item Given the $\hat{q}$ calculated on the previous step, produce confidence intervals or sets in the out-of-sample phase denoted by $C(x_{val})$ with $1-\alpha$ coverage guarantee.
\end{enumerate}

\begin{theorem}[Marginal coverage guarantee]
Suppose that $D=\{(x_i,y_i)\}_{i=1}^{N}$ are i.i.d. samples, if we construct $C(x_{val})$ as indicated above, the following inequality holds for any \textit{non-conformity score function} s and any $\alpha \in (0,1)$
\begin{equation}
\label{theorem1}
    \mathbb{P}(y_{val} \in C(x_{val})) \ge 1-\alpha .
\end{equation}
\end{theorem}

In other words, ICP ensures that the correct value (regression) or class (classification) is within the conformal interval or set with at least $1-\alpha$ confidence, respectively. Readers interested in the proof of (\ref{theorem1}) are referred to \cite{2}.

Vladimir Vovk \cite{4} demonstrated that the distribution of coverage follows the following distribution
\begin{equation}
 \label{betadist}
    \mathbb{E}[\bm{1}\{y_{val} \in C(x_{val})\}| \{(x_i,y_i)\}_{i=1}^{n_{cal}}] \sim  Beta(n_{cal}+1-l,l),
\end{equation}

where $l=\left \lfloor (n_{cal}+1)(1-\alpha)\right \rfloor$.
This result is important to assess whether we are correctly applying ICP. For this purpose Angelopoulos and Bates \cite{3} recommend producing T trials with new calibration sets and then calculate the empirical coverage on T validation sets as
\begin{equation}
    \label{Trials}
    C_j=\frac{1}{n_{val}} \sum_{i=1}^{n_{val}} \bm{1}\{y_{val} \in C(x_{val})\},\qquad j \in \{1,2,...,T\}.
\end{equation}

Thereafter, the histogram of $\{(C_j)\}_{j=1}^{T}$ should be close to a $Beta(n_{cal}+1-l,l)$ and the mean of all trials, $\hat{C_j}$, should be close to $\frac{n_{cal}+1-l}{n_{cal}+1}$, i.e., the mean of the theoretical distribution. 

Although (\ref{theorem1}) is true regardless of the choice of the \textit{non-conformity score function}, the choice of this s function is key, having direct impact on the prediction set sizes. A wide interval (regression) or a big set of classes (classification) is not informative at all. Furthermore, the out-of-sample ML model accuracy is also directly implied in the prediction interval size (regression) and the set size of classes (classification). A ML model model A that is more uncertain, on average, than a ML model B, tends to deliver wider sets, on average, to guarantee $1- \alpha$ coverage.

The previous paragraph intuitively lead to what several authors refer to as adaptiveness. \cite{3}, i.e., a good ICP procedure should deliver small sets on easy inputs and larger sets where the model is uncertain or the input is in fact hard.
Ideally, we seek what is usually called as conditional coverage guarantee \cite{4} given by

\begin{equation}
    \label{condcoverage}
        \mathbb{P}(y_{val} \in C(x_{val})|X=x_{val}) \ge 1-\alpha.
\end{equation}

This is a stronger assumption that ICP does not guarantee; however, there are many heuristic approaches to approximate it \cite{3}.

In the next sections we are going to showcase some practical examples of ICP on regression and classification. Therefore, we strongly advise the reader to follow the remainder of the paper accompanied by \href{https://github.com/Quilograma/ConformalPredictionTutorial/blob/main/Conformal\%20Prediction.ipynb}{this Jupyter Notebook}.

\section{Examples on regression}
In the next examples, for reproducible reasons we use the  \href{https://scikit-learn.org/stable/modules/generated/sklearn.datasets.load_boston.html}{Boston housing prices} dataset, publicly available on sklearn \cite{5}. This dataset has the form $D=\{(x_i,y_i)\}_{i=1}^{506}$, where $x_i \in \mathbb{R}^{13}$ and $y_i \in \mathbb{R}$. In plain words, we have 13 numerical features with which we want to predict the response variable (house price).

\subsection{Naive method}
\begin{figure}[H]
    \centering
    \includegraphics[scale=0.3]{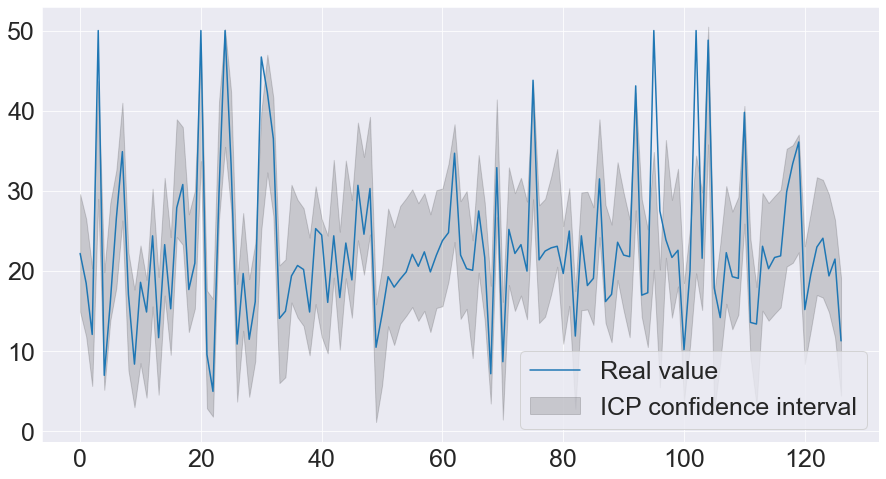}
    \caption{90\% confidence intervals produced with naive ICP regression method.}
    \label{confintervals}
\end{figure}

\begin{figure}[H]%
    \centering
    \subfloat[\centering]{{\includegraphics[scale=0.2]{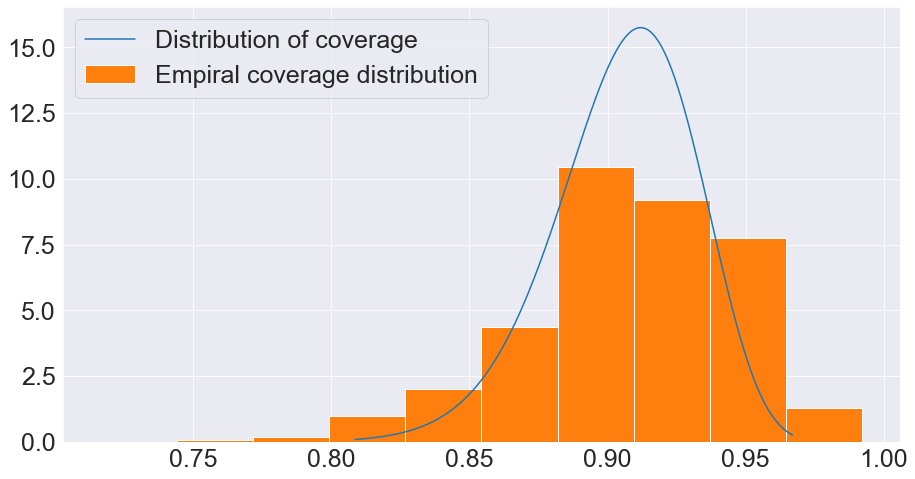} }}%
    \qquad
    \subfloat[\centering]{{\includegraphics[scale=0.2]{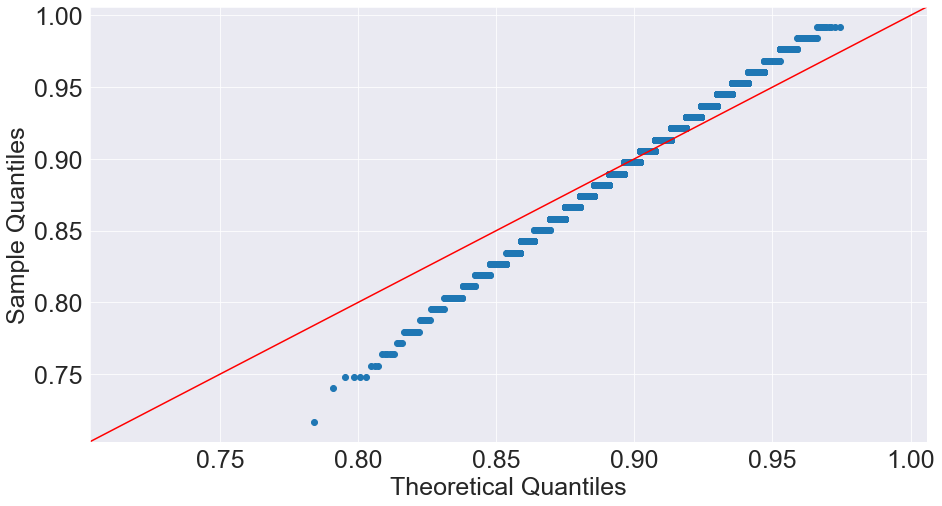} }}%
    \caption{Histogram density vs theoretical beta distribution given by (\ref{betadist}) with T=10000 (panel left). QQPlot of empirical distribution with T=10000 vs theoretical beta distribution (\ref{betadist}) (panel right).}%
    \label{graphs}%
\end{figure}

We start with a naive method, choosing $s(x,y) = |f(x)-y|$ as the \textit{non-conformity score function}. f(x) denotes the ML model forecast on input $x\in \mathbb{R}^{13}$, whereas $y\in \mathbb{R}$ represents the actual value. We randomly split the dataset in 3 mutually exclusive sets as $D=D_{train} \cup D_{cal} \cup D_{val}$. $D_{train}$ contains 50\% of the examples and the others 25\% each. We start by fitting our model on $D_{train}$ and then compute $\hat{q}$ on $D_{cal}$ using $\alpha=0.1$. Recall that  $\hat{q}$ is computed as the $\frac{\left \lceil (n_{cal}+1)(1-\alpha)\right \rceil}{n_{cal}}$ quantile of the \textit{non-conformity scores} $\{(s_i)\}_{i=1}^{n_{cal}}$, i.e., absolute errors in this case. Thereafter, we produce intervals on the validation set as $[f(x_i)-\hat{q},f(x_i)+\hat{q}]$ for each $i \in \{1,2,...,n_{val}\}$  and assess marginal coverage guarantee. We chose a KNN (K-Nearest Neighbors) regressor with 5 neighbors, however, any other ML model could be applied since ICP is model agnostic. We got a $\hat{q}=7.34$ and a marginal coverage of 92.9\%. Fig.(\ref{graphs}) graphically assesses the correctness of this ICP procedure with a total of T=10000 trials using (\ref{betadist}) and (\ref{Trials}). Alternatively, in a statistician manner, we could use the Kolmogorov-Smirnov two sample test to test whether the two distributions are identical \cite{6}. In this example, the empirical coverage sample mean is 90.62\%, close to the mean coverage of 90.55\% indicated by the theoretical distribution (\ref{betadist}). Such a small deviation is not significant. This indicates that, in fact, ICP was correctly applied. Notwithstanding, Fig.(\ref{confintervals}) demonstrates the major drawback of this approach.  The confidence intervals have always the same dimension, equal to $2\hat{q}=14.68$ in this example, regardless of the input, and therefore this naive method has lack of adaptiveness. The next strategies we provide are more sophisticated and try to cope with this shortcoming. They attempt to better approximate the conditional coverage guarantee (\ref{condcoverage}).

\subsection{Adaptive intervals}
In this section, we introduce two methods: (i) conformalized residual fitting (CRF) and (ii) conformalized quantile regression (CQR) \cite{7}. They have the advantage of taking the input into account while ensuring marginal coverage (\ref{theorem1}).

In CRF, we need an additional calibration set so that we split our dataset in four $D=D_{train} \cup D_{cal1} \cup D_{cal2} \cup D_{val}$. We begin by fitting a model $f$ on the training set $D_{train}$, then we fit a second model $r$ on the absolute residuals of $D_{cal1}$ produced by model $f$, i.e., we train $r$ on the set $\{(x_i,|y-f(x_i)|)\}_{i=1}^{n_{cal1}}$. Note that if model $r$ was perfect, then $[f(x_{val})-r(x_{val}),f(x_{val})+r(x_{val})]$ would provide 100\% coverage on the validation set. However, in general, this is not achieved in practice. Indeed, as you can see in our notebook, this resulted only in 66.9\% coverage. Consequently, if we desire 90\% coverage, we need to refine this approach by applying ICP. To do so, we can utilize the following \textit{non-conformity score function}
\begin{equation}
\label{nonconfboost}
    s(x,y)=\frac{|y-f(x)|}{r(x)}.
\end{equation}
Thereafter, given a user-defined level $\alpha$, we compute $\hat{q}$ as the $\frac{\left \lceil (n_{cal2}+1)(1-\alpha)\right \rceil}{n_{cal2}}$ quantile of the \textit{non-conformity scores} on the second calibration set $D_{cal2}$ and form $1-\alpha$ confidence intervals as 
\begin{equation}
    [f(x_{val})-\hat{q} r(x_{val}),f(x_{val})+\hat{q}r(x_{val})].
\end{equation}
 In our experiment for $\alpha=0.1$ we got $\hat{q}=2.97$ with 93.7\% coverage on the validation set.

Now, we introduce CQR. Quantile regression (QR) is a long-lasting result in the literature \cite{8,9}. It attempts to learn the $ \epsilon \in (0,1)$ quantile based on the conditional distribution $Y=y|X=x$. QR can be applied on the top of any ML model by simply changing the loss function to the \textit{quantile loss} function, also known as \textit{pinball loss} due to its resemblance to a pinball ball movement. This loss function can be mathematically expressed as
\begin{equation}
    L_{\epsilon}(y,f(x)|)=\max\left( \epsilon(y-f(x)), ( (\epsilon-1)(y-f(x))\right).
\end{equation}
Therefore, if we pretend 90\% coverage with QR we could fit a ML model using $\epsilon_1=0.05$ and $\epsilon_2=0.95$. Nevertheless, these quantiles are only estimates of the true quantiles and thus the interval $[\hat{t}_{\epsilon_1}(x_{val}),\hat{t}_{\epsilon_2}(x_{val})]$ might not verify (\ref{theorem1}). Actually, in our experiments, QR provided 69.29\% validation coverage even though we asked for 90\%. As a consequence, we should use the following \textit{non-conformity score function} s on the calibration set
\begin{equation}
    s(x,y)=(\hat{t}_{\frac{\alpha}{2}}(x)-y,y-\hat{t}_{1-\frac{\alpha}{2}}(x)).
\end{equation}
After computing $\hat{q}$ as usual, we are then in conditions of producing intervals with \textit{validity}  (\ref{theorem1}) as
\begin{equation}
    [\hat{t}_{\frac{\alpha}{2}}(x_{val})-\hat{q},\hat{t}_{1-\frac{\alpha}{2}}(x_{val})+\hat{q}]
\end{equation}
Note that if $\hat{q}$ is positive, then the interval gets wider. On the contrary, if $\hat{q}$ is negative, it shrinks. After calibrating with ICP we got $\hat{q}=3.03$ and 92.1\% validation coverage.

\begin{figure}[H]%
    \centering
    \subfloat[\centering]{{\includegraphics[scale=0.2]{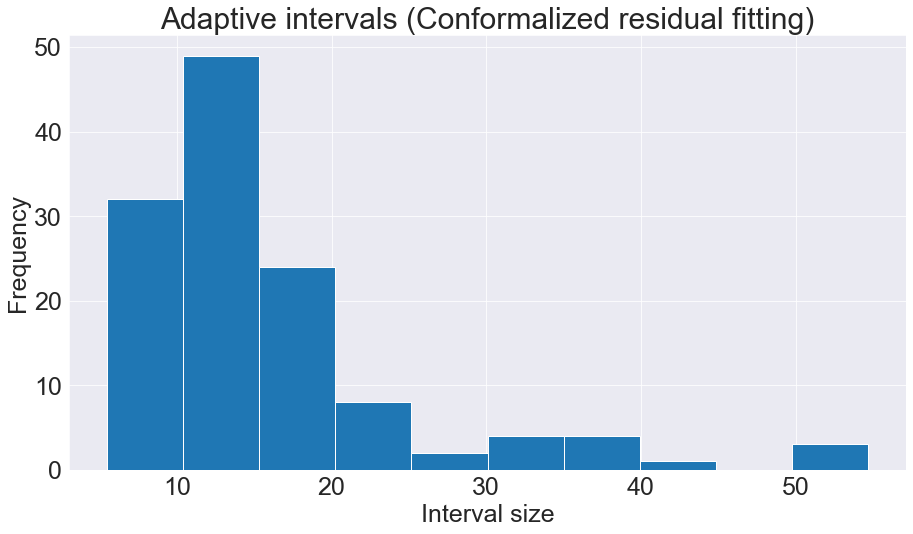} }}%
    \qquad
    \subfloat[\centering]{{\includegraphics[scale=0.2]{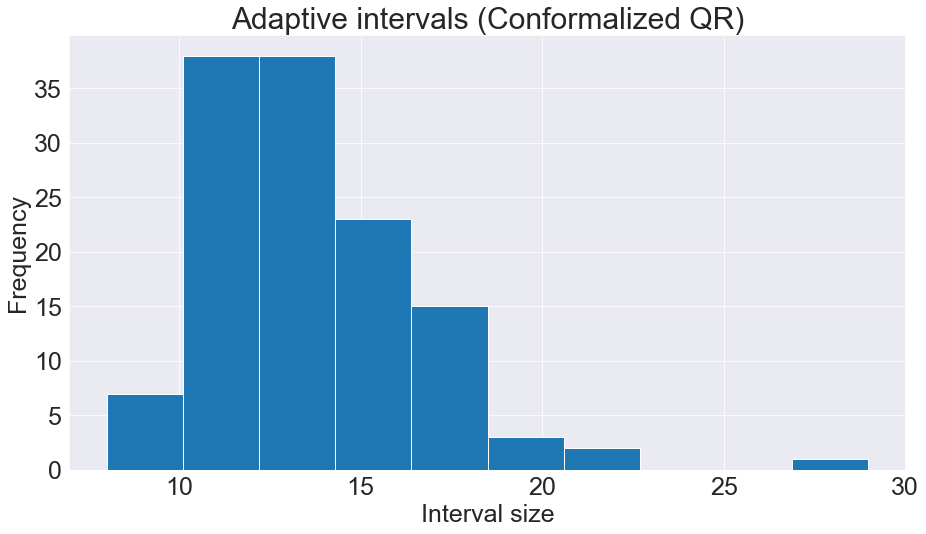} }}%
    \caption{Adaptive interval sizes validation distribution for CRF (panel left) and CQR (panel right). CRQ provided better results, i.e., greater validation coverage with smaller intervals.}%
\end{figure}

\begin{figure}[H]
    \centering
    \includegraphics[scale=0.4]{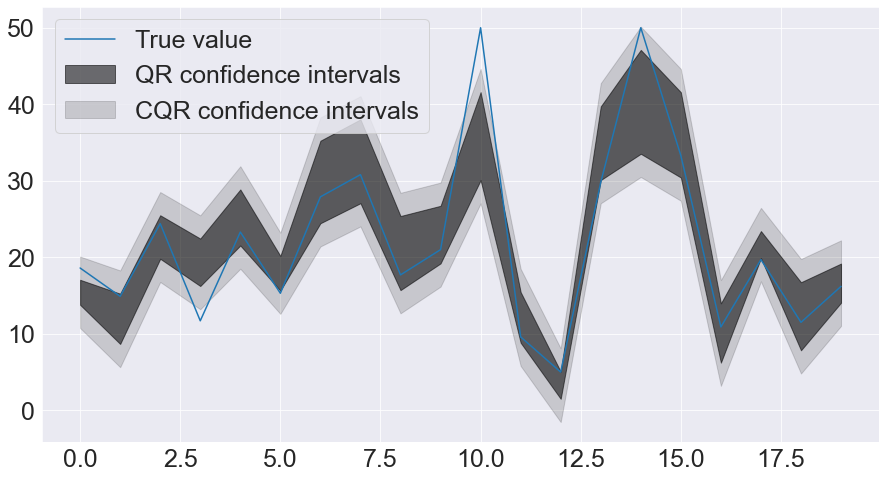}
    \caption{Confidence intervals with QR and CQR after a $\hat{q}$ band shift.}
\end{figure}

\section{Examples on classification}
In our classification example we will use a built-in keras dataset \href{https://www.tensorflow.org/api_docs/python/tf/keras/datasets/cifar10/load_data}{cifar10}. It is a dataset of 60.000 images with a total of 10 classes. More precisely, $D=\{(x_i,y_i)\}_{i=1}^{60000}$, where $x_i\in \mathbb{R}^{32 \times 32 \times 3}$ (RGB image) and $y_i \in\{0,1,...,9\}$ (10 classes).

\subsection{Naive method}
We start by illustrating an application on classification with a intuitive method, afterwards we present other strategies that are known to be more adaptive.

Basically, in a multi classification problem, our model attempts to approximate the following distribution $\mathbb{P}(Y=y_i|X=x_i)$. In our example it would mean: given an image $x_i$ what is the probability of it belonging to class $y_i$? To attain such an objective, we have utilized a convolutional neural network (CNN) with a softmax output layer, yet there are multiple different ways to achieve this. Mathematically, our trained model can be represented by the following function $f:\mathbb{R}^{32 \times 32 \times 3} \rightarrow (0,1)^{10}$, i.e., given a image $x_i$ we output a set of 10 probabilities that all together sum to 1. We will represent $f(x)_{y_{true}}$ as the softmax output of the true class. First, we need to split the dataset in three mutually exclusive sets as usual $D=D_{train} \cup D_{cal} \cup D_{val}$. Second, we train our ML model on $D_{train}$. Third, we choose $s(x,y_{true})=1-f(x)_{y_{true}}$  as the \textit{non-conformity score function} and get a list of scores $s_1,s_2,...,s_{n_{cal}}$ on the calibration set $D_{cal}$. Fourth, we compute $\hat{q}$ as the $\frac{\left \lceil (n_{cal}+1)(1-\alpha)\right \rceil}{n_{cal}}$ quantile of the aforementioned scores. Finally, we are now ready to get prediction sets as $C(x_{val})=\{y: f(x_{val})_y \geq 1- \hat{q}\}$ for every $x_{val}$ in the validation set. Regarding our example for $\alpha=0.1$ the outcome was $\hat{q}=0.96$ and $88.18\%$ validation coverage. 

\begin{figure}[H]%
    \centering
    \subfloat[\centering]{{\includegraphics[scale=0.2]{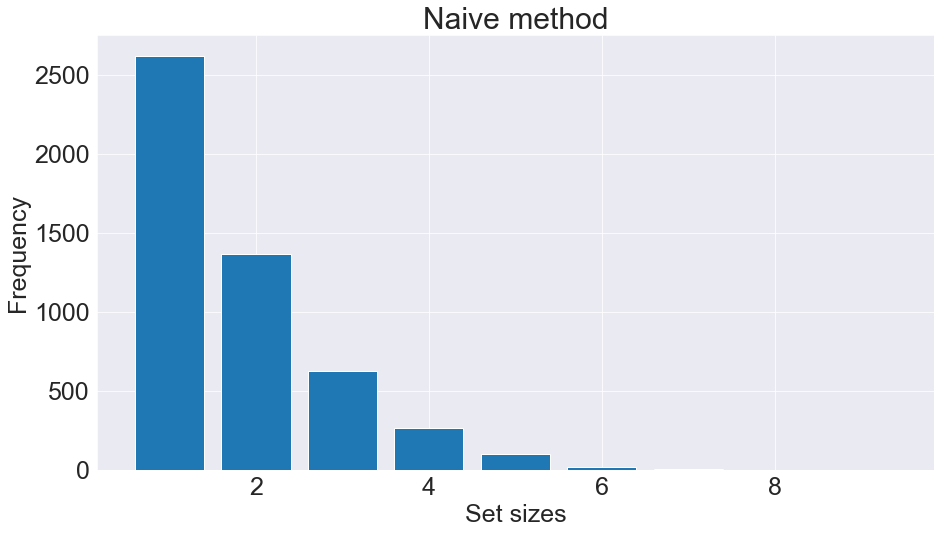} }}%
    \qquad
    \subfloat[\centering]{{\includegraphics[scale=0.2]{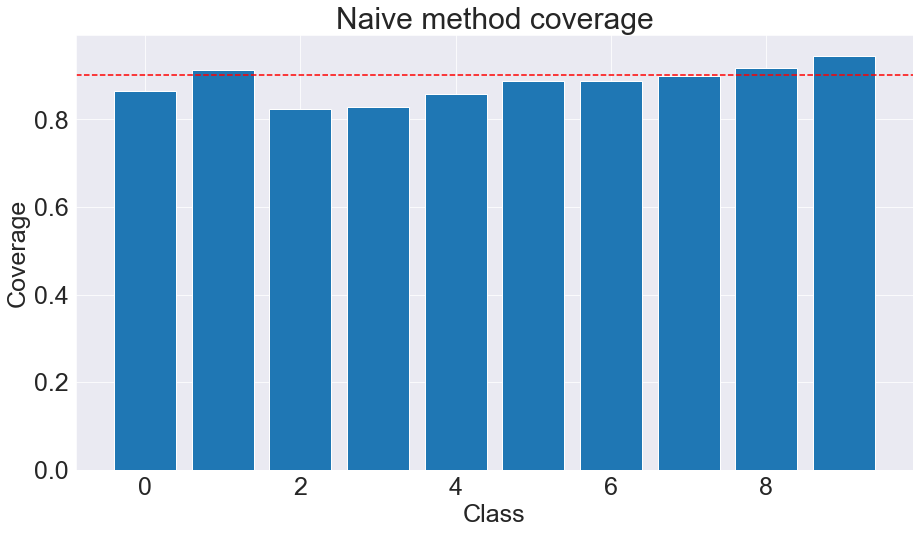} }}%
    \caption{(Naive method) Set size produced in our classification example (panel left). Coverage by class (panel right)}%
    \label{naivecmethod}
\end{figure}

\subsection{Class-balanced}
The former algorithm often undercovers some classes and overcover others as seen in Fig.(\ref{naivecmethod}). It guarantees (\ref{theorem1}) but it might provide high coverages on some classes and lower on others.
The algorithm we are about to introduce guarantees the following
\begin{equation}
    \mathbb{P}(y_{val}\in C(x_{val})|Y=y_{val})\ge 1-\alpha.
\end{equation}

\begin{figure}[H]%
    \centering
    \subfloat[\centering]{{\includegraphics[scale=0.2]{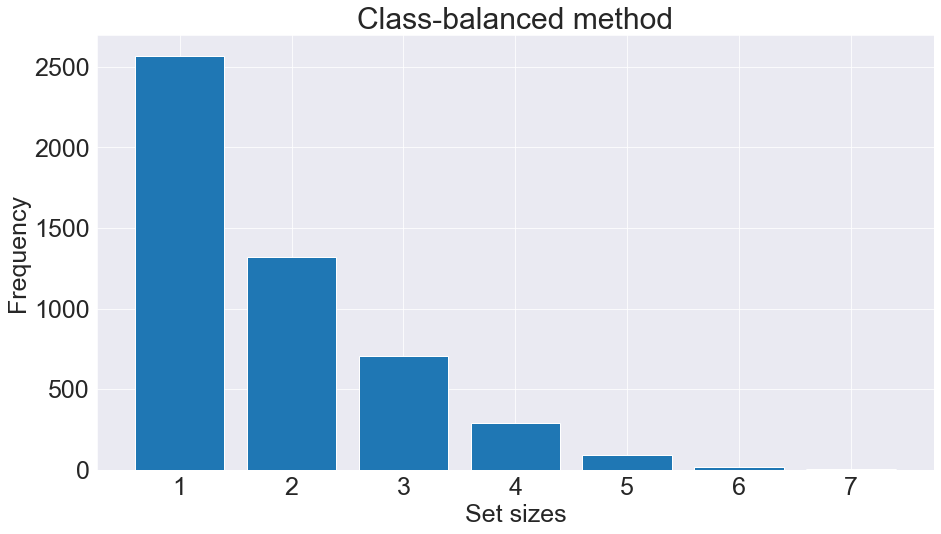} }}%
    \qquad
    \subfloat[\centering]{{\includegraphics[scale=0.2]{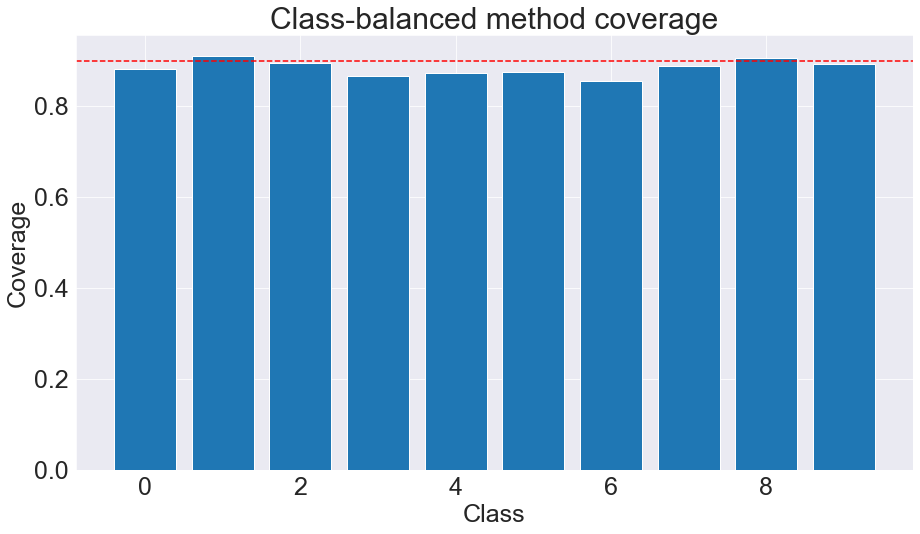} }}%
    \caption{(Class-balanced method) Set size produced in our classification example (panel left). Coverage by class (panel right)}%
    \label{classbalancedmethod}
\end{figure}

The philosophy of the class-balanced method is the same as the naive method, however, we calculate a different $\hat{q}$ for each class, i.e., we stratify by class. In terms of our example this means that we have have a list $\hat{q}^{(0)},\hat{q}^{(1)},...,\hat{q}^{(9)}$, one for each class. Prediction sets are then computed as $C(x_{val})=\{y: f(x_{val})_y \geq 1- \hat{q}^{(y)}\}$. Fig.(\ref{classbalancedmethod}) shows that this strategy achieves in fact class balanced results despite small deviations due to the finite sample issue.

\subsection{Adaptive prediction sets}
\begin{figure}[H]%
    \centering
    \subfloat[\centering]{{\includegraphics[scale=0.2]{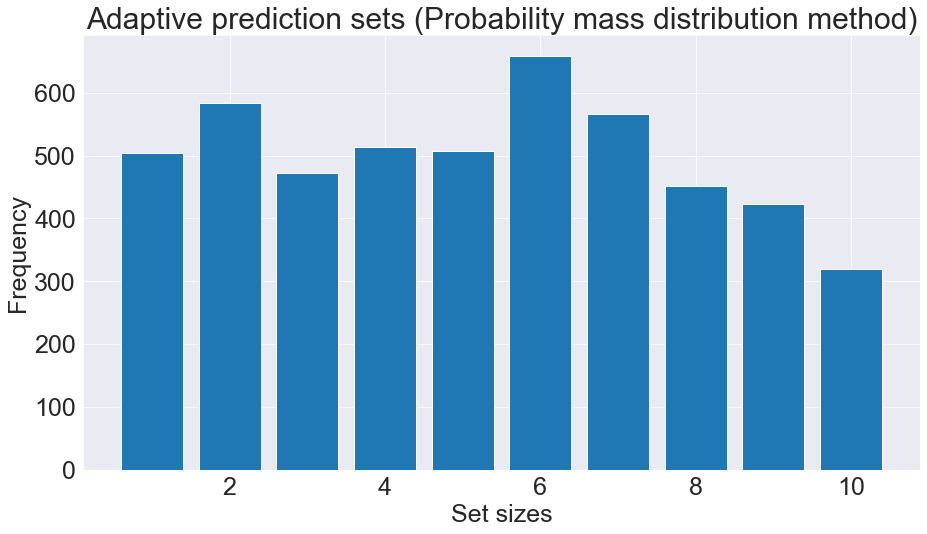} }}%
    \qquad
    \subfloat[\centering]{{\includegraphics[scale=0.2]{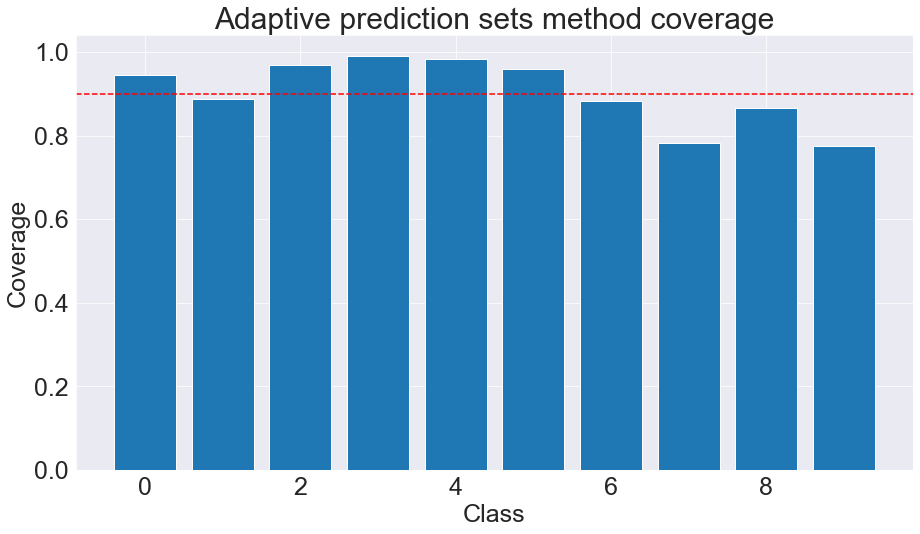} }}%
    \caption{(Adaptive prediction sets method) Set size produced in our classification example (panel left). Coverage by class (panel right)}%
    \label{APSmethod}
\end{figure}

Thus far, we have only utilized the softmax output of the true class and therefore we miss important information about the other classes. This gap lead to a method that is often called adaptive prediction sets (APS) \cite{11,12}. We need some additional notation before showing how APS words. We denote $\pi(x)$ as a permutation of $f(x)$ that orders the softmax output in descending order, i.e., from the most likely class to the less likely. The \textit{non-conformity score
function} of APS is as follows
\begin{equation}
    s(x,y)=\sum_{i=1}^k\pi(x)_y,
\end{equation}
where k is the minimum number of classes we have to go through until we reach the true class. In plain words, s delivers the cumulative density we need to include until we reach the true class. Next, we compute $\hat{q}$ as usual and compute confidence sets as $C(x_{val})=\{y:\sum_{i=1}^k\pi(x)_y \ge \hat{q}\}$. In plain words, we form prediction sets by including every class, reversely sorted considering the softmax output, until we exceed $\hat{q}$ cumulative density.
Again, in our example, given $\alpha=0.1$ we got $\hat{q}=0.999995$ and 90.44\% validation coverage. In spite of being more adaptive, sets produced by APS tend to be larger as seen in Fig. (\ref{APSmethod}).

\section{Conclusion}
 Our brief introduction finishes here. We hope that this hands-on introduction has convinced the reader to use ICP in practice and motivated to further investigate about advanced conformal prediction strategies in more complex scenarios. Distribution-free uncertainty quantification is still in its infancy, despite the upward tendency seen in recent years.

%Bibliography
\bibliographystyle{unsrt}  
\bibliography{references}

\end{document}